%% file: paper.tex
\documentclass[10pt,conference,a4paper]{IEEEtran}
\usepackage{times}
\usepackage{epsfig}
\usepackage{graphicx}
\usepackage{amsmath}
\usepackage{amssymb}
\usepackage{multirow}
\usepackage[listofformat=parens, justification=centering, subrefformat=subparens]{subfig}
\usepackage[pagebackref=true,breaklinks=true,letterpaper=true,colorlinks,bookmarks=false]{hyperref}
\renewcommand\cap[3]{\caption[#2]{\label{#1}\textsc{#2}. \small\textit{#3}}}
\graphicspath{{./}}

\DeclareMathOperator{\sign}{sign}

\ifCLASSINFOpdf
\else
\fi
\begin{document}
%
\title{Are Facial Attributes Adversarially Robust?}


\author{\IEEEauthorblockN{Andras Rozsa, Manuel G\"unther, Ethan M. Rudd, and Terrance E. Boult }
\IEEEauthorblockA{University of Colorado at Colorado Springs\\
Vision and Security Technology (VAST) Lab\\
Email: http://vast.uccs.edu/contact-us
}}



%


\maketitle

\begin{abstract}
Facial attributes are emerging soft biometrics that have the potential to reject non-matches, for example, based on mismatching gender.
To be usable in stand-alone systems, facial attributes must be extracted from images automatically and reliably.
In this paper, we propose a simple yet effective solution for automatic facial attribute extraction by training a deep convolutional neural network (DCNN) for each facial attribute separately, without using any pre-training or dataset augmentation, and we obtain new state-of-the-art facial attribute classification results on the CelebA benchmark.
To test the stability of the networks, we generated adversarial images -- formed by adding imperceptible non-random perturbations to original inputs which result in classification errors -- via a novel fast flipping attribute (FFA) technique.
We show that FFA generates more adversarial examples than other related algorithms, and that DCNNs for certain attributes are generally robust to adversarial inputs, while DCNNs for other attributes are not. 
This result is surprising because no DCNNs tested to date have exhibited robustness to adversarial images without explicit augmentation in the training procedure to account for adversarial examples.
Finally, we introduce the concept of natural adversarial samples, i.e., images that are misclassified but can be easily turned into correctly classified images by applying small perturbations. 
We demonstrate that natural adversarial samples commonly occur, even within the training set, and show that many of these images remain misclassified even with additional training epochs.
This phenomenon is surprising because correcting the misclassification, particularly when guided by training data, should require only a small adjustment to the DCNN parameters.
\end{abstract}

%
\IEEEpeerreviewmaketitle

\input{introduction}
\input{attributes}
\input{adversarials}
\input{conclusion}

\section*{Acknowledgment}
This research is based upon work supported in part funded in part by NSF IIS-1320956 and in part by the Office of the Director of National Intelligence (ODNI), Intelligence Advanced Research Projects Activity (IARPA), via IARPA R\&D Contract No. 2014-14071600012. The views and conclusions contained herein are those of the authors and should not be interpreted as necessarily representing the official policies or endorsements, either expressed or implied, of the ODNI, IARPA, or the U.S. Government. The U.S. Government is authorized to reproduce and distribute reprints for Governmental purposes notwithstanding any copyright annotation thereon.



%

%
%
%

{\small
\bibliographystyle{ieee}
\bibliography{paper}
}

\end{document}

%% file: introduction.tex
\section{Introduction}

Facial attributes have several interesting properties from a recognition perspective. First, they are semantically meaningful to humans, which offers a level of interpretation beyond that achieved by most conventional recognition algorithms. This allows for novel applications, including descriptive searches (e.g., ``Caucasian female with blond hair'')~\cite{kumar2008facetracer,kumar2011describable,scheirer2012multi}, verification systems~\cite{kumar2009attribute}, facial ordering~\cite{parikh2011interactively}, social sentiment analysis~\cite{zhang2015learning}, and demographic profiling. Second, they provide information that is more or less independent of that distilled by conventional recognition algorithms, potentially allowing for the creation of more accurate and robust systems, narrowing down the search space, and increasing efficiency at match time. Finally, facial attributes are interesting due to their ability to convey meaningful identity information about a previously unseen face, e.g., not enrolled in a gallery or used to train a classifier.

\begin{figure}[!t]
  \centering
  \subfloat[\label{fig:ffa:a}Fixing a Natural Adversarial on Gender: from \emph{female} to \emph{male}]{\includegraphics[width=\columnwidth]{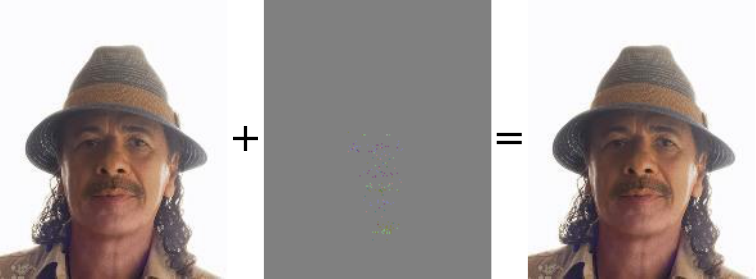}}\\
  \subfloat[\label{fig:ffa:b}Flipping Age: from \emph{young} to \emph{old}]{\includegraphics[width=\columnwidth]{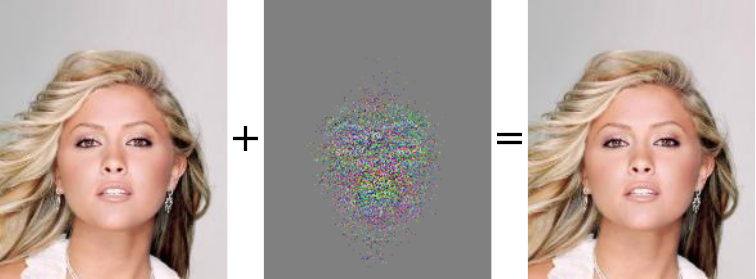}}

\cap{fig:ffa}{Adversarial Samples by Attribute Flipping}{This paper demonstrates the problem of natural adversarial samples and how to generate adversarial examples for attributes by the novel fast flipping attribute (FFA) technique. Adversarial examples are formed by adding small, non-random perturbations to original inputs: while the modifications can remain imperceptible to human observers, vulnerable machine learning models classify the original and adversarial examples differently. In \protect\subref*{fig:ffa:a}, we show a natural adversarial man (misclassified as a woman) that is ``corrected'' by an imperceptible perturbation. In \protect\subref*{fig:ffa:b}, we show a correctly classified young person flipped to  ``old''. Perturbations are magnified by factor of 50 to enhance visibility.}
\end{figure}

Recent state-of-the-art approaches to facial attribute classification \cite{liu2015deep} have leveraged powerful deep learning models, which derive a generic feature space representation that is optimized for face identification or verification. They then use the truncated network to extract features upon which to train per-attribute binary classifiers. Although this approach leads to a compact representation, it does not explicitly incorporate attribute information into the learnt feature space. While attributes that relate to facial identity may \emph{be implicitly captured} by this representation, there is little evidence to suggest that the representation will effectively distill non identity related attribute information, e.g., \textit{smiling}. On the contrary, intuition suggests that a network trained to discriminate identities would learn to ignore non identity related attributes. Motivated by recent research \cite{levi2015age}, we hypothesize that by explicitly incorporating attribute information into deep learning representations, we can attain superior performance, especially for non identity related attributes. We provide supporting evidence for this hypothesis by advancing the state of the art on the CelebA benchmark -- the largest publicly available attribute dataset -- simply by using separate deep networks, each trained on raw attribute data. Moreover, unlike the former state-of-the-art approaches, we are able to attain such high performance \emph{without any form of data augmentation}.

While our approach is state of the art on the CelebA benchmark, recent research~\cite{szegedy2013intriguing,goodfellow2014explaining} raises the question: \emph{Does using pure end-to-end deep networks, i.e., not simply as feature extractors, but as attribute classifiers themselves, induce a risk of a non-robust attribute representations for real-world applications?} Specifically, Szegedy et al.~\cite{szegedy2013intriguing} discovered that deep neural networks are susceptible to carefully chosen perturbations of even a few pixels. By adding such selected perturbations -- that cannot be perceived by humans -- to the original images, the resulting \emph{adversarial images} become misclassified with high confidence.

Much research on adversarial images has been conducted since, and -- to our knowledge -- all of these images have been easily generated independently of the dataset, network topology, training regime, hyperparameter choice, and activation type.
In our experiments we attempt to generate adversarial images over a random subset of the CelebA dataset \cite{liu2015deep} using the fast gradient sign (FGS) method~\cite{goodfellow2014explaining}, and a new algorithm for introducing adversarial images -- the \emph{fast-flipping attribute} (FFA) algorithm -- that efficiently leverages backpropagation \emph{without requiring groundtruth labels}.
We find that for both FGS and FFA \emph{attribute classifications are difficult to change}, at least for some attributes, and that the number of adversarial images \emph{does not decrease} during training.


To date, adversarial images have been presented as inputs under the presence of slight artificial perturbations where the original input is correctly classified and the adversarial input is misclassified.
In this paper, we pose the reverse question: ``Do there exist misclassified inputs on which we can induce small artificial perturbations to correct the classification?'', or ``Do adversarial images naturally occur?''
We find that the answer is yes, there are images in the training set which, even after they are used for training, are misclassified by given facial attribute network but can be flipped to the proper classification via an imperceptible perturbation.
Further, we find that even with additional training, many of these natural adversarial samples are \emph{not} learnt by the networks.
This is surprising because correctly learning these natural adversarial samples should require only a minor adjustment to the network parameters.

In Fig.~\ref{fig:ffa}, we show two examples of adversarial images that occurred in our experiments.
Fig.~\subref{fig:ffa:a} contains a natural adversarial image of a man (left) that is misclassified as a woman, where a small perturbation (center) applied to the image (right) would correct the classification.
The reverse is shown in Fig.~\subref{fig:ffa:b}, where a correctly classified young person (left) is turned into an old person (right) by adding a small perturbation (center).
Note that the real perturbations are much smaller than shown in Fig.~\ref{fig:ffa}, for visualization we magnified the pixel changes by a factor of 50.
Since perturbations can be positive and negative, gray pixels correspond to no change.

The contributions of this paper are as follows:

\begin{itemize}

\item We \emph{advance the state of the art} on the CelebA attribute classification benchmark with relative reduction in classification error of over 25\,\%.

\item We generate adversarial images for each of the attribute networks and find that our facial attribute networks attain no additional robustness to adversarial images with longer training. 

\item We introduce the notion of \emph{natural adversarial images} and analyze their prevalence using our data and networks. We find that the frequency of naturally occurring adversarial images is quite large, accounting for nearly 73\,\% of the training set images that are incorrectly classified by facial attribute networks.

\item We introduce the \emph{fast flipping attribute} (FFA) algorithm for adversarial image generation and demonstrate that it is successful at flipping attribute classifications.

\end{itemize}

%% file: attributes.tex
\section{Attribute Classification}
\label{sec:attributes}

\begin{figure*}[t!]
  \centering\includegraphics[width=\linewidth]{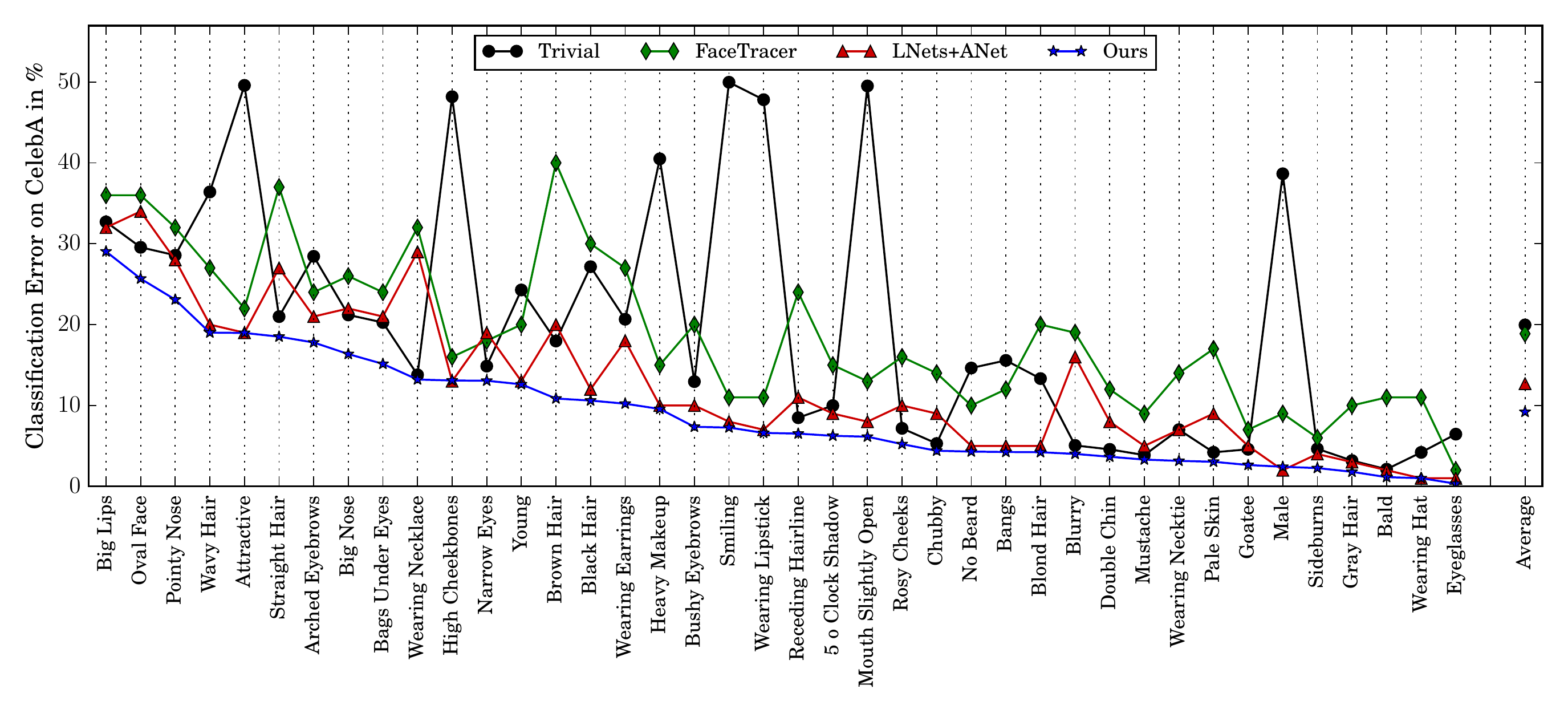}\vspace*{-3ex}
  \cap{fig:ErrorRates}{Attribute Classification Error on CelebA}{This figure shows the classification errors on the test set of the CelebA dataset for Our algorithm compared to three other algorithms, sorted based on the result of Ours. The Trivial approach simply assigns each attribute the score of the majority class. The results of FaceTracer and LNets+ANet are taken from Liu et al.~\cite{liu2015deep}.}
\end{figure*}

The automatic classification of facial attributes was first pioneered by Kumar et al.~\cite{kumar2009attribute}.
Their classifiers depended heavily on face alignment and each attribute used learnt combinations of features from hand-picked facial regions (e.g., cheeks, mouth, etc.).
The feature spaces consisted of various simple normalizations and aggregations of color spaces and image gradients.
Different features were learnt for each attribute, and one RBF-SVM per attribute was independently trained for classification.

More recent approaches leverage deep convolutional neural networks (DCNNs) to extract features.
Liu et al.~\cite{liu2015deep} use three DCNNs -- a combination of two \emph{localization networks} (LNets) and an \emph{attribute recognition network} (ANet) -- to first localize faces and then extract facial attributes.
The ANet is first trained on external data to identify people and then fine-tuned using all attributes to extract features that are fed into independent linear SVMs for the final attribute classification.
This approach is the current state of the art on the CelebA dataset.

Contrary to \cite{liu2015deep} where the feature space is not necessarily attribute derived, our approach uses one DCNN per attribute, trained only on the CelebA training set. We use the outputs of these networks directly as classification results without requiring secondary classifiers.

\subsection{Network Classification}

For our attribute classifiers, we adopted the 16 layer VGG network topology from \cite{parkhi2015deep}, with two modifications.
First, we altered the dimension of the RGB image input layer from $224\times224$ pixels to $178\times 218$ pixels, the resolution of the aligned CelebA images.
Second, we replaced the final softmax layer with Euclidean loss on the label.
We chose Euclidean loss -- as opposed to softmax, sigmoid, or hinge loss -- because attributes lie along a continuous range, while sigmoids tend to enforce saturation and hinge-loss enforces a large margin, neither of which is consistent with our intuition. 

For an image $x$ with label $y \in \{-1,+1\}$ indicating the absence or presence of an attribute, respectively, let $f(x)$ be the DCNN classification decision.
Then the loss $J$ is:
\begin{equation}
J(\theta,x,y) = {|| f(x) - y ||}^{2},
\end{equation}
where $\theta$ are the parameters of the DCNN model.

To maintain comparability with other research reporting on the same dataset, and since the dataset only contains binary attribute labels, we decided to apply a classification function to the network output that was trained with Euclidean loss.
For input $x$, the classification result $c(x)$ and its corresponding error $e(x,y)$ are obtained by thresholding $f(x)$ at 0:

\begin{equation}
  \label{eq:accuracy}
  c(x) \! = \! \begin{cases} \! +1 & \! \text{if~} f(x) \! > \! 0 \\ \! -1 & \! \text{otherwise,} \end{cases} \quad
  e(x,y) \! = \! \begin{cases} 0 & \! \text{if~} y \! \cdot \! c(x) \! > \! 0 \\ 1 & \! \text{otherwise.} \end{cases}
\end{equation}

The classification error over the whole dataset $X$ of $N$ images with attribute labels $Y$ is then given by:
\begin{equation}
  \label{eq:classification_error}
  E(X,Y) = \frac1N\sum\limits_{n=1}^N \bigl(e(X_n,Y_n)\bigr)\,.
\end{equation}

\subsection{Experiments}

We conducted a comparison of our separate per-attribute neural networks with other attribute algorithms on the CelebA dataset \cite{liu2015deep}.
CelebA consists of more than 200K images which show faces in a variety of different facial expressions, occlusions and illuminations, and poses from frontal to full profile.
Approximately 160K images are used for training, and the remaining 40K images are equally split up into validation and test sets.
Each image is annotated with binary labels of 40 facial attributes.
We conducted our evaluation using the set of pre-cropped face images included in the dataset, which are aligned using hand-annotated key-points.

Due to memory limitations, we set the training batch size to 64 images per training iteration. Thus, the training requires approximately 2500 iterations to run a full epoch on the training set.
In opposition to \cite{parkhi2015deep}, we did not incorporate any dataset augmentation or mirroring; we trained networks purely on the aligned images.
We selected a learning rate of $10^{-5}$.
During training, we updated DCNN weights using an \textit{RMSProp} update rule with an inverse learning rate decay policy.
Using the GPU implementation of Caffe \cite{jia2014caffe}, we trained all 40 networks until convergence on the validation set, which occurred between two and ten epochs depending on the attribute.

A comparison of our results on the CelebA test set with the original FaceTracer approach by Kumar et al.~\cite{kumar2009attribute} as well as the LNets+ANet state-of-the-art approach of Liu et al.~\cite{liu2015deep} is shown in Fig.~\ref{fig:ErrorRates}.
Due to the highly biased distributions of attribute labels in the CelebA dataset, we also included a \textit{Trivial} algorithm, which simply predicts the class with the higher occurrence in the training set.
For some attributes such as Attractive or Male (which are approximately balanced in the test set), the Trivial classifier obtains high errors, while for attributes like Narrow Eyes or Double Chin, the Trivial classifier even outperforms the previous state-of-the-art approach.
Our networks are at least able to outperform the Trivial approach for all attributes, which is not true for the other two approaches.

Our approach yields a mean classification error of 9.20\,\%, a relative improvement of 27.5\,\% over the state-of-the-art (12.70\,\% classification error) and 51\,\% improvement over the FaceTracer system (18.88\,\% classification error).
Interestingly, we are ``only'' 54\,\% better (in terms of relative improvement) than the Trivial system, which obtains 19.96\,\% mean classification error.
For certain attributes, especially those not related to face identity (e.g., Wearing Necklace, Wearing Earrings, Blurry), our approach dramatically advances the state of the art.
LNets+ANet outperforms our approach only for a few attributes, but never by more than a percentage point in classification error.

%% file: adversarials.tex
\section{Adversarial Images for Attributes}
\label{sec:adversarials}

\begin{figure*}
  \centering\includegraphics[width=\linewidth]{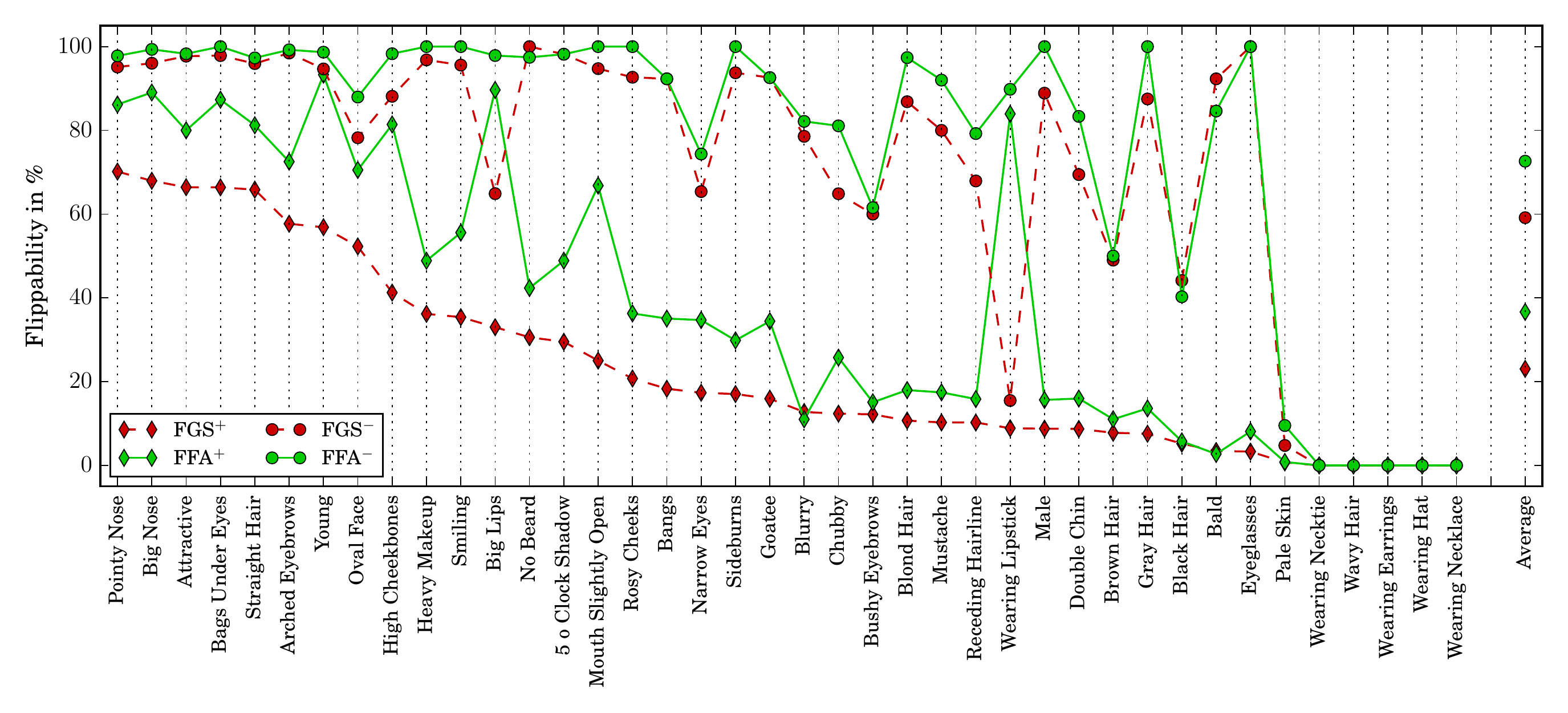}\vspace*{-3ex}
  \cap{fig:Flippable}{Adversarial Success}{This figure shows the success rates of FGS and FFA adversarial generation for adversarial examples (correctly classified images, $^+$) and natural adversarial samples (incorrectly classified images, $^-$) generated on the converged networks trained with Euclidean loss, sorted by success rate of FGS$^+$ and based on 1000 images randomly selected from the CelebA training set.}
\end{figure*}

An adversarial image is an image which looks very close to (and is generally indistinguishable from) an original image from the perspective of a human observer, but differs dramatically in classification by a machine learnt classifier.
Multiple techniques have been proposed to create adversarial examples.
The first reliable technique \cite{szegedy2013intriguing} uses a box-constrained optimization (L-BFGS).
Starting with a randomly chosen modification, it aims to find the smallest perturbation in the input space that causes the perturbed image to be classified as a predefined target label.
Baluja et al.~\cite{baluja2015virtues} proposed generating affine perturbations, applying them to input samples, and then observing how models respond to these perturbed images.
While, from an adversarial perspective, this approach has the advantage of not requiring internal network representations, it relies on ``guess and check'', i.e., it creates random perturbations and determines if the results are misclassified, which can be prohibitively expensive.

Goodfellow et al.~\cite{goodfellow2014explaining} introduced a more efficient algorithm to produce adversarial perturbations.
Their fast gradient sign (FGS) method creates perturbations by using the sign of the gradient of loss with respect to the input.
FGS is more efficient than L-BFGS because it requires only one gradient which can be effectively calculated via backpropagation, while L-BFGS needs multiple.
Experiments demonstrate that FGS reliably causes a wide variety of learning models to misclassify their perturbed inputs, including ``shallow'' models, but that deep networks are especially susceptible  \cite{goodfellow2014explaining}.
Note that FGS is based directly upon network information. The gradient of loss defines the direction, while a line-search is used to determine the magnitude necessary to make the perturbed input adversarial with minimal deviation from the original.

Although the definitions of \emph{adversarial example} vary \cite{szegedy2013intriguing, goodfellow2014explaining, baluja2015virtues}, at their core, adversarial examples are modified inputs formed by {\em imperceptible non-random perturbations} that are misclassified by machine learning models.
Hence, humans should not even perceive differences between adversarial and original inputs.
To  formalize the definition, let $x$ be an input image \emph{correctly classified} as $y$.
An adversarial perturbation $\eta$ is given if the perturbed image $\tilde{x}=x+{\eta}$ is not classified as $y$:
\begin{align}
f(x) &= y &\mathrm{and} && f(\tilde x) &\neq y.
\end{align}
This is a necessary but not a sufficient condition as the modification needs to be \emph{imperceptible}.
Various measures such as $L_1$, $L_2$, or $L_\infty$ distances have been used to show how close the perturbed images are to their originals. However, these measures are not well matched to human perception \cite{sabour2015adversarial} as they are very sensitive to even small geometric distortions that may still result in plausible images.
Instead, we seek a measure of similarity in a psychophysical sense.
The \textit{perceptual adversarial similarity score} (PASS) \cite{Rozsa_2016_CVPR_Workshops} measures similarity $S(\tilde x,x)$ based on structural photometric-invariant differences, by first performing a homography alignment to maximize the \textit{enhanced correlation coefficient} (ECC) \cite{evangelidis2008parametric} between the perturbed image and the original, then computing the \textit{structural similarity index} (SSIM) \cite{flynn2013image} between the aligned images.
The homography transform removes differences due to plausible looking translations and rotations, while SSIM tends to measure only structural differences not stemming from perturbations that appear as plausible photometric differences.
We believe that the resultant PASS score is a more suitable measure of the degree to which an image is adversarial.
Consistent with \cite{flynn2013image}, we adopt the PASS threshold $\tau = 0.95$ as a cutoff for \textit{adversarial}.

\subsection{Adversarial Image Generation}

We explore two approaches for generating the necessary perturbations $\eta$.
Goodfellow et al.~\cite{goodfellow2014explaining} introduced the fast gradient sign (FGS) method to find adversarial perturbations.
Given an image $x$, FGS searches for perturbations causing mislabeling using the sign of the gradient of loss:
\begin{equation}
  \label{eq:fgs}
  \eta_{_{ \mathrm{fgs}}} = w\,\sign\bigl(\nabla_{x}J(\theta,x,y)\bigr)\,.
\end{equation}

FGS takes steps in the direction that is defined by the gradient of loss in order to cause mislabeling.
This requires knowledge of the label $y$ of the image $x$.
We introduce a novel approach -- the fast flipping attribute (FFA) algorithm -- that directly relies on the (binary) classification scores.
We postulate that inverting the classification score and calculating the gradient of the inverted score with respect to the input image will eventually provide a direction where adversarial perturbations can be found.
Formally:
\begin{equation}
  \label{eq:ffa}
  \eta_{_{\mathrm{ffa}}} = - w\,\frac{\partial f(x)}{\partial x}
\end{equation}
can be obtained by backpropagating the inverted classification score from the decision layer, i.e., the layer that calculates $f(x)$.

By using FGS or FFA, we can obtain directions that define varying perturbations $\eta$ with respect to a given input image $x$.
To effectively search along those directions for the smallest perturbations that cause classification errors, we apply a line-search technique with increasing step sizes to reach the weight $w$ in Eqns. \eqref{eq:fgs} and \eqref{eq:ffa} that causes mislabeling. By doubling the step size after each step, those directions can be quickly discovered. When the line-search oversteps and the classification of the perturbed example changes, we apply a binary search within the latest section of the line-search to find the smallest possible adversarial perturbation. Due to these enhancements, adversarial generation with FFA and FGS approaches achieve comparable computational efficiency. The generated adversarial images have rounded discrete pixel values in range $[0,255]$.



\subsection{Natural Adversarial Images}

As of today, adversarial images have been artificially generated via a computational process, but no one -- to our knowledge -- has yet addressed whether adversarial inputs occur among natural images: are there misclassified images for which infinitesimal changes to inputs yield correct classifications?
If so, this has tremendous ramifications on the sensitivity and robustness of decision boundaries.
We seek to explore how often adversarial images naturally occur.
Thus, we formalize the novel concept of \textit{natural adversarial images}.
Let $x'$ be an \emph{incorrectly classified} image whose correct label is $y$.
Then $x'$ is a natural adversarial image if there exists a perturbation $\eta'$ such that perturbed image $\tilde x'=x'+\eta'$ is \emph{correctly classified} as $y$:
\begin{equation}
  f(x') \neq  y \qquad\mathrm{and}\qquad f(\tilde x') = y,
\end{equation}
under the premise that $\mathcal S(\tilde x',x') < \tau$.
Hence, a natural adversarial sample is an image that is misclassified, but that will be correctly classified when an imperceptible modification is applied.
Interestingly, the same processes used to generate adversarial images can be used to analyze if a misclassified input is a natural adversarial.

\subsection{Experiments}

To test and compare adversarial generation with FGS and FFA, we randomly selected 1000 images of the CelebA training set and performed experiments trying to flip attributes.
For each attribute and both the correctly and incorrectly classified images, we counted the number of times in which an adversarial image could be created, i.e., where an $\eta$ exists for which $\mathcal S(x, x+\eta) < \tau$, for an $\eta$ generated by either of the two algorithms.
The results per attribute can be obtained in Fig.~\ref{fig:Flippable}.
Interestingly, for some attributes such as Big Nose or Young, most input images can be turned adversarial, while for others like Wavy Hair or Wearing Necklace, adversarial samples cannot be formed at all.
Even more astonishingly, incorrectly classified images can be turned into adversarial examples more often than correctly classified images.
Also, the number of images, for which we could generate adversarial images using FFA is generally higher than for FGS, where almost all images that spawned FGS adversarial images also spawned FFA adversarial images.

In an attempt to test in which stage of the network training more adversarial images exist, we also tried to generate adversarial images for the same 1000 examples on DCNNs that were trained for two epochs.
Intuition would suggest that DCNNs are able to learn adversarial samples that exist at two epochs, and the total number of adversarial images would decrease.
Especially for natural adversarial images, i.e., images that were misclassified by DCNNs at two epochs for which imperceptible modifications to those images would make them correct, we assume that further training would make the networks learn these examples.
The results of these experiments, which are listed in detail in the top half of Tab.~\ref{tab:Adversarials}, are counter-intuitive.
For FGS, the total number of adversarial images over all attributes that we were able to create slightly increased from 8827 to 10587, while more than half of the images (5884) were in both sets (which includes images that were misclassified before and now classified correctly and vice versa).
For FFA, the numbers for DCNNs at two epochs and after convergence are similar, and for most vulnerable input images we could create adversarial samples on both converged and unconverged networks.

In general, the total number of incorrectly classified images for which adversarial samples could be created was consistent between two epochs and convergence, though at least half of the original input images differed.
This also means that half of the images, for which an imperceptible modification would have been sufficient to classify them correctly, were \emph{not} corrected with additional training.
At least we found that of the 1279 images that were natural FFA adversarial samples at two epochs and that are not natural adversarial samples on the converged networks, the majority (1048) were classified correctly by the converged networks. The numbers for FGS are similar.

\begin{table}[t!]
  \renewcommand{\arraystretch}{1.15}
  \centering\vspace*{2.0ex}
  \begin{tabular}{l|c|c|c|c}
  	\hline
    Loss							& Adv. Type		& Two Epochs 	& Converged	& Overlap\\
    \hline\hline
	\multirow{4}{*}{Euclidean}	& FGS$^+$		& 6393			& 8345		& 4616 \\
								& FGS$^-$		& 2434			& 2242		& 1268 \\
								& FFA$^+$		& 13621			& 13268		& 9881 \\
								& FFA$^-$		& 2918			& 2753		& 1639 \\
	\hline
	\multirow{4}{*}{Softmax}		& FGS$^+$		& 6833			& 12047		& 5896 \\
								& FGS$^-$		& 2385			& 2431		& 1315 \\
								& FFA$^+$		& 16371			& 21218		& 15259 \\
								& FFA$^-$		& 2844			& 2573		& 1721 \\
  	\hline

  \end{tabular}
  \cap{tab:Adversarials}{Adversarial Images}{This table shows the numbers of FGS and FFA adversarial images generated on models at two epochs and after convergence, trained with either Euclidean or softmax loss. We show numbers for adversarial examples (correctly classified images, $^+$) and natural adversarial samples (incorrectly classified images, $^-$) over all attributes. The overlap represents the number of images that are sources of adversarial instances both at two epochs and after convergence.}
\end{table}\vspace*{0.0ex}

%% file: conclusion.tex
\section{Discussion and Conclusion}
\label{sec:conclusion}

In this paper we employed a simple and effective method to train deep convolutional neural networks (DCNNs) to perform binary facial attribute classification.
Experiments on the CelebA dataset show that with 9.20\,\% average classification error our approach outperforms the current state of the art (12.70\,\%). This performance gain is statistically significant, resulting in $p < 10^{-27}$ in a paired T-test.
Afterward, we introduced the fast flipping attribute (FFA) algorithm, a fast and robust method to generate adversarial images by flipping the binary decision of the DCNN.
We demonstrated that FFA can create more adversarial examples than the related fast gradient sign (FGS) method, but it has the limitation -- by design -- that it can only be applied to binary classification networks.

In Sec.~\ref{sec:adversarials}, we demonstrated that a greater number of training epochs does not make DCNNs more robust to adversarial images.
On the other hand, at two epochs DCNNs obtained an average classification error of 9.78\,\% that is statistically significantly ($p < 10^{-10}$ in the paired T-test) higher than the 9.20\,\% that we obtained with the converged DCNNs.

\begin{figure*}
  \centering
  \includegraphics[width=\textwidth]{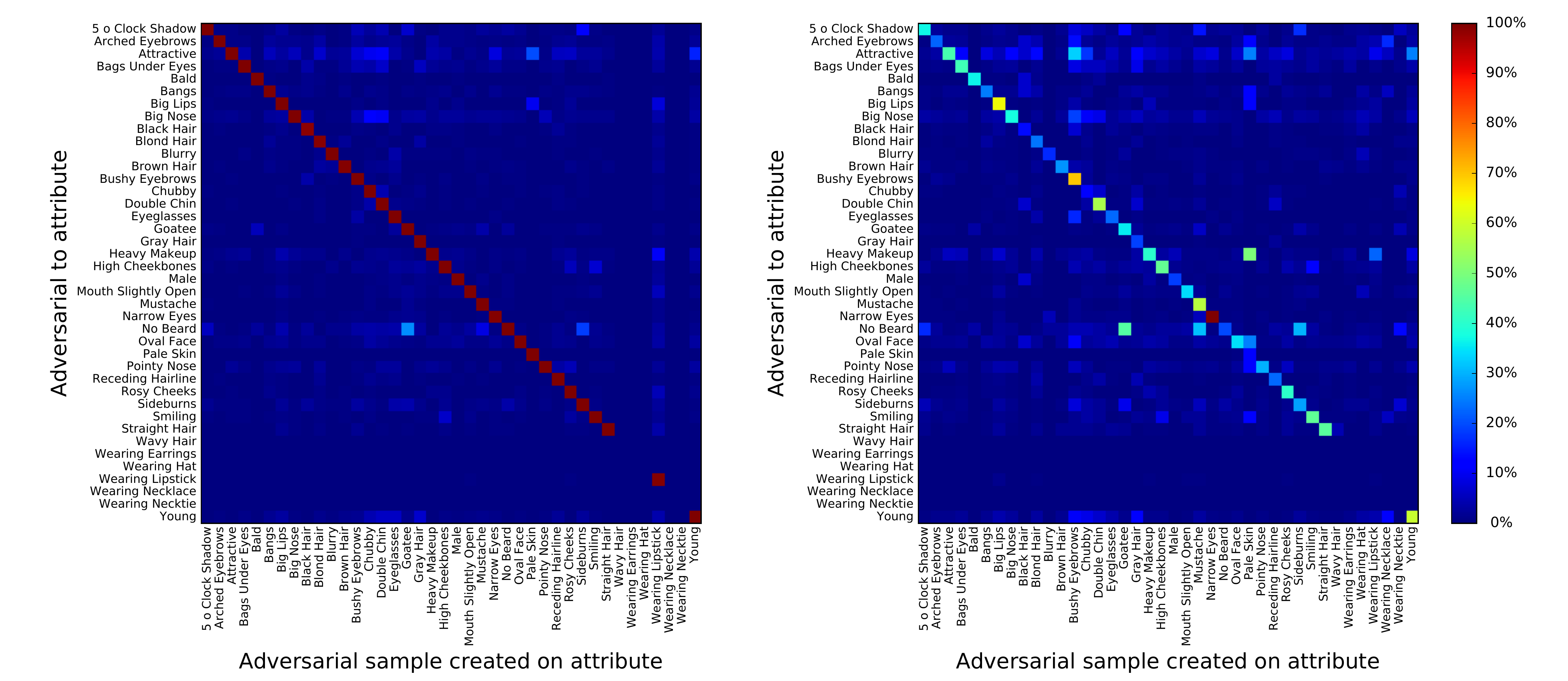}\vspace*{0.0ex}
  \cap{fig:Portability}{Adversarial Portability}{This figure shows the adversarial portability of attributes throughout networks. On the left, FFA adversarial images created on the converged networks of a given attribute are tested on all converged networks. On the right, FFA adversarial images created on networks after two epochs are tested on the fully converged networks. All networks were trained using Euclidean loss.}
\end{figure*}

A natural question is, to what extent do adversarial inputs generated on one facial attribute network affect classifications by other networks. To test this, we tried to create adversarial examples on a network for one attribute and evaluated whether these perturbed inputs influence other attributes as well. The results are shown in Fig.~\ref{fig:Portability}.
For these experiments, we used the same 1000 images and report the average percentage of cross-attribute portability.

On the left, we present the portability of adversarial samples between the converged networks.
Interestingly, most of the perturbations for one attribute do not influence other attributes.
However, for some attributes we can observe higher correlations.
For example, 26\,\% of the adversarial samples for Goatee also flip No Beard, and adversarial samples created for several attributes like Chubby, Double Chin, Pale Skin, and Young influence the Attractive attribute.
For certain attributes, like Wearing Earrings and Wearing Hat, we were not able to create adversarial samples; hence, these columns are empty.

On the right of Fig.~\ref{fig:Portability}, we report the results of a similar experiment.
We created adversarial examples on the two epoch networks and checked if they were still adversarial on the converged networks.
On the diagonal, we can see how many adversarial inputs created on the two epoch networks remained adversarial on the converged DCNNs. Note that the Narrow Eyes network had already converged after two epochs.

On average, more than 30\,\% of the adversarial images were not learnt by the networks, i.e., the same small perturbations still caused misclassifications.
Notably, the cross-attribute portability of adversarial samples even increased, e.g., around 50\,\% of the adversarial images created on the Pale Skin network at two epochs flipped the classification of the converged Heavy Makeup network.

In Sec.~\ref{sec:attributes}, we chose an uncommon loss function --  Euclidean loss -- to train our DCNNs.
A key rationale for choosing this loss function is that in parallel attribute research~\cite{moonECCV2016} we found that an extension of the Euclidean loss function was the most trivial way to incorporate multi-label domain adaptive loss, making single-label Euclidean loss networks a natural point of comparison for that work.
Readers might ask which loss function leads to better classification performance: Euclidean or softmax?
To answer this question, we conducted experiments, training attribute classification networks using softmax loss.
Compared to Euclidean loss networks, DCNNs trained with softmax loss ended up with a slightly higher classification error of 9.30\,\%, which is not statistically significant ($p \approx 0.0003$ in a paired T-test).
Also, we found that DCNNs trained with softmax loss are more vulnerable to adversarial examples.
As detailed in Tab.~\ref{tab:Adversarials}, using the same 1000 examples from the CelebA training set (cf. Sec.~\ref{sec:adversarials}) we can generate 16021 adversarial samples (over all 40 attributes) using FFA on the converged DCNNs trained with Euclidean loss, and 23791 using FFA on softmax loss DCNNs.
This is consistent with other research which demonstrates that DCNNs trained with softmax loss are generally vulnerable to adversarial examples.

Since all attributes leverage a shared representation in a multi-label network~\cite{moonECCV2016}, flipping one attribute label has the potential side effect of changing more labels of other correlated attributes. We leave further analysis of mixed-objective adversarial generation to future work.
In order to further improve the overall performance of our networks, future work can also consider fine-tuning trained DCNN models with adversarial samples or hard positives~\cite{Rozsa_2016_CVPR_Workshops}, corrected natural adversarial samples, or even augmented natural adversarial samples containing perturbations with lower magnitudes than otherwise needed for correcting misclassifications.
